\documentclass{article}
\usepackage{spconf,amsmath,graphicx,hyperref}
\usepackage{booktabs}
\usepackage{multirow}

\title{Exploring Audio Hallucination in Egocentric Video Understanding}
\name{\begin{tabular}{c}
Ashish Seth$^{1,2}$, Xinhao Mei$^{1}$, Changsheng Zhao$^{1}$, Varun Nagaraja$^{1}$, Ernie Chang$^{1}$, Gregory P. Meyer$^{1}$,\\
Gael Le Lan$^{1}$, Yunyang Xiong$^{1}$, Vikas Chandra$^{1}$, Yangyang Shi$^{1}$, Dinesh Manocha$^{2}$, Zhipeng Cai$^{1}$
\end{tabular}}

\address{$^{1}$Meta, $^{2}$University Of Maryland, College Park}
\begin{document}
\maketitle
\begin{abstract}
Egocentric videos provide a distinctive setting in which sound serves as crucial cues to understand user activities and surroundings, particularly when visual information is unstable or occluded due to continuous camera movement. State--of--the--art large audio--visual language models (AV--LLMs) can generate multimodal descriptions. However, we show in this work that they are prone to audio hallucinations, often inferring sounds from visual cues that are visible but not heard. We present a systematic and automatic evaluation framework for analyzing audio hallucinations in egocentric video through a targeted question-answering (Q/A) protocol. We curate a dataset of 300 egocentric videos and design 1,000 sound-focused questions to probe model outputs. To characterize hallucinations, we propose a grounded taxonomy that distinguishes between foreground action sounds from the user activities and background ambient sounds. Our evaluation shows that advanced AV--LLMs, such as Qwen2.5 Omni, exhibit high hallucination rates, achieving only 27.3\% and 39.5\% accuracy on Q/As related to foreground and background sounds, respectively. With this work, we highlight the need to measure the reliability of multimodal responses, emphasizing that robust evaluation of hallucinations is essential to develop reliable AV--LLMs. Project page: \url{https://cs20s030.github.io/EgoAVU/}
\end{abstract}
\section{Introduction}
\label{sec:intro}
\begin{figure}[t]
    \centering
    \includegraphics[width=1.0\linewidth]{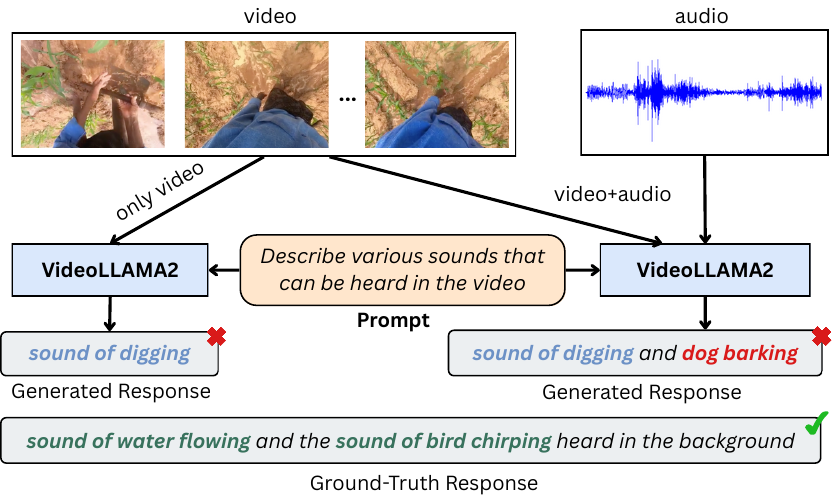}
    \caption{\small \textbf{Audio hallucinations in AV-LLMs.} For example, VideoLLAMA2~\cite{cheng2024videollama} incorrectly infers the foreground sound ``sound of digging'' (shown in blue) from the presence of digging tools, in the video-only (no audio) inference path. Importantly, this visual bias persists even when audio is provided: in the video+audio path, the model continues to rely on the visual cue, hallucinating the foreground sound ``digging'' and additionally hallucinating the background sound ``dog barking'' (shown in red).}
    \label{fig:hero_diag}
\end{figure}

Audio–visual language models (AV-LLMs) and multimodal large language models (MLLMs) have demonstrated impressive abilities to generate rich descriptions from both videos and audios, supporting tasks such as event understanding, action recognition, and sound description~\cite{yin2024survey, zhu2021deep}. Yet, despite their multimodal design, these models often hallucinate. Prior work has largely examined visual hallucinations (e.g., recognizing objects or actions not actually present)~\cite{bai2024hallucination, seth2025egoillusion}, but audio hallucinations, where sound descriptions conflict with the actual audio and instead follow visual cues, remain underexplored, particularly in first-person (egocentric) video~\cite{grauman2022ego4d, damen2018scaling}. We argue that studying audio hallucinations in egocentric settings is especially important, since frequent camera motions and occlusions degrade visual inputs, while audios provide reliable cues for understanding the wearer activities and surroundings~\cite{chen2024action2sound}. To investigate the cause of such hallucinations, we analyze the behavior of AV-LLMs, as illustrated in Fig.\ref{fig:hero_diag}: when asked to describe sounds with and without audio, state-of-the-art AV-LLMs such as VideoLLAMA2\cite{cheng2024videollama} produce nearly identical responses, underscoring their strong reliance on visual context. These visually driven hallucinations in generated audio descriptions reveal a key limitation: \textit{current AV-LLMs overweight visual features and fail to accurately ground audio content}.


\begin{figure*}[t]
    \centering
    \includegraphics[width=1.0\linewidth]{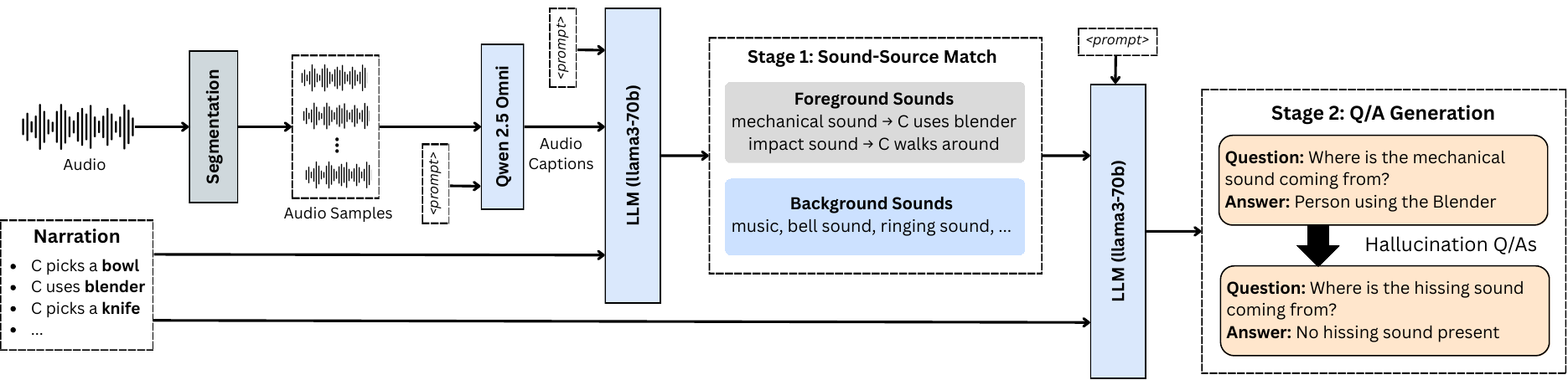}
    \caption{\small \textbf{An illustration of the data curation pipeline.} Egocentric videos are first segmented into 10-second audio clips. These clips are then captioned using Qwen2.5 Omni~\cite{xu2025qwen2}. In Stage 1 (Sound-Source Match), an LLM (LLaMA3-70B~\cite{dubey2024llama}) associates these audio captions with narrations from the video to match sounds with their sources, distinguishing between foreground and background noises. Finally, in Stage 2 (Q/A Generation), the same LLM generates question-answer pairs based on the matched sound sources, including pairs designed to test for "hallucinations" by asking about sounds that are not present.}
    \label{fig:datacurration}
\end{figure*}

\noindent{\textbf{Main Contribution.}} In this work, we present a systematic framework to analyze audio hallucinations in egocentric video. We curate a dataset of \textit{300 egocentric videos} and design \textit{1,000 sound-focused questions} to probe model outputs. To categorize hallucinations, we propose a grounded taxonomy specific to egocentric settings, focusing on two prominent sound types: foreground action sounds (from the wearer’s activities) and background ambient sounds (from the environment). Our evaluation shows that state-of-the-art AV-LLMs, such as Qwen2.5 Omni~\cite{xu2025qwen2}, exhibit high hallucination rates, achieving only \textit{27.3\%} and \textit{39.5\%} accuracy on Q/As related to foreground and background sounds, respectively.



\section{Related Work}
\label{sec:format}

\noindent{\textbf{Audio-Visual Language Models (AV-LLMs).}} Recent advances in AV-LLMs and multimodal large language models (MLLMs) have enabled rich multimodal understanding, including video captioning, sound description, and action recognition~\cite{yin2024survey, zhu2021deep}. These models integrate visual and auditory modalities to generate coherent descriptions; however, they often struggle to accurately ground audio content, particularly in complex or noisy environments.

\noindent{\textbf{Egocentric Video Understanding.}} Egocentric video presents unique challenges due to frequent camera motion, occlusions, and the mixture of foreground and ambient sounds~\cite{grauman2022ego4d, damen2018scaling, seth2025egoillusion}. Despite extensive work on egocentric action recognition and multimodal reasoning, audio hallucination remains largely understudied in first-person settings. Our work addresses this by proposing a source-grounded taxonomy and systematic evaluation of audio hallucinations for egocentric video.

\section{Methodology}
\label{sec:pagestyle}
\subsection{Dataset Curation}
To systematically probe the influence of visual context on audio perception in egocentric videos, we introduce a curated dataset derived from 300 clips of the Ego4D corpus~\cite{grauman2022ego4d}, encompassing 50 distinct visual scenarios. As shown in Fig~\ref{fig:datacurration}, our data curation pipeline, starts with the segmentation of the source audio into non-overlapping 10-second clips. For each segment, we generate preliminary audio annotations using the Qwen2.5 Omni model~\cite{xu2025qwen2} to produce descriptive captions of sound events. These machine-generated captions, along with time-aligned human narrations detailing object interactions, are then processed by the LLaMA3-70B~\cite{dubey2024llama} model in a two-stage refinement process. In Stage 1 (Sound-Source Match), the model first categorizes sounds as either foreground or background events and then correlates the foreground sounds with their originating actions specified in the narrations (e.g., associating a ``mechanical sound" with ``uses blender"). Subsequently, in Stage 2 (Q/A Generation), these validated sound-source pairs are utilized to generate question-answering pairs. We generate both factual Q/A pairs concerning present audio events and carefully crafted ``hallucination Q/As'' that query plausible but absent sounds. 

\subsection{Stage 1: Sound-Source Match}
\begin{figure}[t]
    \centering
    \includegraphics[width=1.0\linewidth]{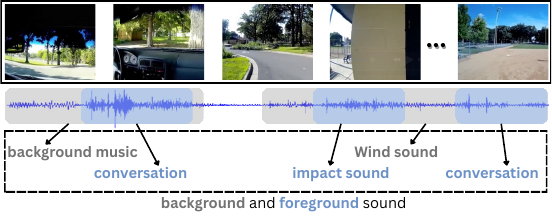}
    \caption{\small \textbf{Illustration of various background and foreground sounds present in a typical egocentric video.} Foreground sounds are often produced by the person’s actions, for example, an “impact sound” generated while walking, whereas background sounds, such as “wind noise,” originate from the surrounding environment and are independent of the person’s direct activities.}
    \label{fig:foreground_background}
\end{figure}

The primary objective of Stage 1 is to accurately ground detected audio events within the egocentric context by distinguishing between foreground and background sounds and identifying the specific source of foreground events~\cite{chen2024action2sound}. As illustrated in Fig~\ref{fig:foreground_background}, egocentric videos inherently capture a complex acoustic environment, comprising sounds directly linked to the person's interactions and those originating from the broader environment. To address this, we leverage the LLaMA3-70B~\cite{dubey2024llama} model, fed with both the Qwen2.5 Omni~\cite{xu2025qwen2} generated audio captions and the human-annotated egocentric narrations (which specify the person's time-aligned actions and interacted objects).
Through specific prompting, the LLM first categorizes each sound event into two types:
\begin{itemize}
\item \textbf{Foreground Sounds:} These are audio events directly attributable to the actions of egocentric device users or their interaction with objects, thereby possessing the visual evidence within the video. We design specific prompts so the LLM can identify these by matching descriptions in the audio captions (e.g., ``mechanical sound," ``impact sound") with corresponding actions or objects detailed in the narration (e.g., ``C uses blender," ``C walks around").
\item \textbf{Background Sounds:} These contains ambient audio events that are not directly caused by the user actions or object interactions. The LLM categorizes sounds as background when no direct grounding evidence is found in the provided narration for the specific temporal segment. For example, environmental sounds such as ``music," ``bell sound," or ``ringing sound" that emanate from the surrounding environment rather than the direct engagement from the user.
\end{itemize}

\subsection{Stage 2: Q/A Generation}
Following the grounding of audio events in Stage 1, the second stage of our pipeline focuses on generating a diagnostic question-answering benchmark to systematically evaluate a model's audio reasoning and its susceptibility to visual biases. Using the sound-source associations learned in Stage 1, we prompt the LLaMA3-70B~\cite{dubey2024llama} model to create two categories of question-answer pairs: Factual and Hallucinated. 

First, we generate \textbf{Factual Q/A pairs} that directly query the information derived from Stage 1. These questions probe the source of a verified foreground sound (e.g., Question: ``Where is the mechanical sound coming from?"; Answer: ``Person using the Blender") or confirm the presence of an identified background sound. These pairs establish a baseline for the ability of the model to report on genuine audio events. Second, we generate \textbf{Hallucination Q/A pairs} designed to induce and detect visually-grounded auditory hallucinations. We prompt the model to formulate questions about sound events that are plausible within the egocentric scene but are absent from the audio track. For example, in a kitchen scene where no frying occurs, the model might generate: Question: ``Where is the hissing sound coming from?". The ground-truth answer for such queries explicitly states the sound's absence (e.g., ``No hissing sound present"). This framework enables us to directly measure a model’s reliance on visual priors over true auditory understanding. To ensure the quality and diversity of our benchmark, the generated Q/A pairs are manually reviewed to filter out redundant and poorly-formed questions. This curation process results in a final dataset of 1,000 Q/A pairs, balanced between open-ended questions that ask for the source of a sound (``Where is the sound of $<sound>$ coming from?") and close-ended questions requiring a Yes/No response.

\section{Experimental Setup}
\noindent{\textbf{Baseline Details.}} We evaluate four state-of-the-art AV-LLMs that jointly process audio and visual information, encompassing diverse architectures and training paradigms. Our evaluation includes ImageBind-LLM~\cite{han2023imagebind}, VideoLLaMA2~\cite{cheng2024videollama}, Qwen2.5-Omni~\cite{xu2025qwen2} with 7B active parameters, and MiniCPM~\cite{hu2024minicpm} with 8B active parameters. Notably, while ImageBind-LLM and MiniCPM were not trained on egocentric video, both VideoLLaMA2 and Qwen2.5-Omni incorporate egocentric video data.   

\begin{figure}
    \centering
    \includegraphics[width=1.0\linewidth]{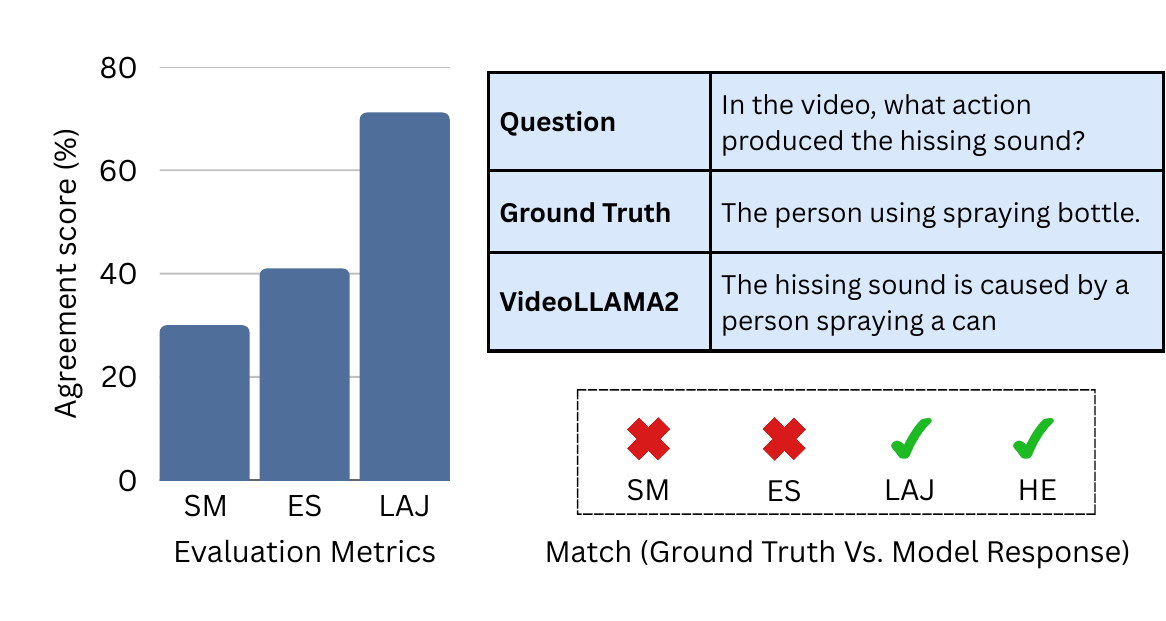}
    \caption{\small \textbf{Agreement scores between automatic metrics and human evaluation (HE).} We compare three automatic evaluation metrics: \textit{string matching} (SM), \textit{embedding similarity} (ES), and \textit{LLM-as-judge} (LAJ) on Qwen2.5 Omni~\cite{xu2025qwen2} responses. Among these, LAJ achieves the highest agreement with human evaluation.}
    \label{fig:compare_eval}
\end{figure}

\noindent{\textbf{Evaluation.}} Similar to prior works ~\cite{seth2025egoillusion, plizzari2024egptempo}, we evaluate three automatic scoring schemes, \textit{string matching} with regular expressions, \textit{embedding-based similarity} using NV-Embed-v2~\cite{lee2024nv}, and \textit{LLM-as-judge}~\cite{gu2024survey} using GPT-4o~\cite{hurst2024gpt}, by comparing their predictions against human judgments on a randomly selected set of 200 Q/A pairs. Agreement with human evaluation is measured using an agreement score, defined as the percentage of responses where the human evaluation and the chosen metric produce the same result. As shown in Fig.~\ref{fig:compare_eval}, the LLM-as-judge approach achieves the highest alignment with human evaluation. In contrast, string matching and embedding similarity perform significantly worse, especially on stronger models such as \emph{Qwen2.5-Omni}~\cite{xu2025qwen2}, which generate longer and more detailed responses than models like \emph{VideoLLaMA2}~\cite{cheng2024videollama}. Therefore, we apply LLM-as-judge for automatic evaluation in later larger scale experiments.

\section{Results}

\subsection{Quantitative Results}
Table~\ref{tab:qa_results} presents the performance of four AV-LLMs on factual (Fact.) and hallucinatory (Hal.) question answering (Q/A) across both foreground and background sounds in egocentric videos. While most AV-LLMs perform close to random guessing, SOTA models such as Qwen2.5 Omni reach 56.2\% on foreground factual and 63.4\% on background factual Q/A, but only 27.3\% and 39.5\% on the corresponding hallucinatory tasks, highlighting substantial room for improvement. We observe that VideoLLaMA2 and MiniCPM perform comparably, both surpassing ImageBind-LLM, with the largest gains on background sounds (e.g., 52.6\% for VideoLLaMA2 compared to 44.5\% for ImageBind-LLM). From these results, we identify two systematic trends. First, across all models and evaluation settings, background sounds consistently yield higher accuracy than foreground sounds. For instance, we find that MiniCPM improves from 39.4\% on foreground factual Q/A to 41.7\% on background factual Q/A, while Qwen2.5 Omni shows a similar increase from 56.2\% to 63.4\%. \textit{This trend suggests that AV-LLMs exploit visual cues rather than grounding sounds in the video when answering questions about foreground sounds.} Second, we find that factual Q/A consistently outperforms hallucinatory Q/A across all models, \textit{underscoring the difficulty of hallucination detection.} The gap is particularly noticeable in ImageBind-LLM, which drops from 28.5\% factual to 21.2\% hallucinatory Q/A on foreground sounds.

\begin{table}[t]
\centering
\resizebox{\columnwidth}{!}{%
\begin{tabular}{lcccc}
\toprule
\multirow{2}{*}{\textbf{Models}} & \multicolumn{2}{c}{\textbf{Foreground Sounds}} & \multicolumn{2}{c}{\textbf{Background Sounds}} \\
\cmidrule(lr){2-3} \cmidrule(lr){4-5}
 & \textbf{Fact. Q/A} & \textbf{Hal. Q/A} & \textbf{Fact. Q/A} & \textbf{Hal. Q/A} \\
 & (Acc.\%) & (Acc.\%) & (Acc.\%) & (Acc.\%)\\
\midrule
ImageBind LLM~\cite{han2023imagebind}   & 28.5\% & 21.2\% & 44.5\% & 20.5\% \\
MiniCPM~\cite{hu2024minicpm}         & 39.4\% & 19.3\% & 41.7\% & 29.6\% \\
VideoLLaMA2~\cite{cheng2024videollama}    & 35.2\% & 20.7\% & 52.6\% & 30.4\% \\
Qwen2.5 Omni~\cite{xu2025qwen2}   & \textbf{56.2\%} & \textbf{27.3\%} & \textbf{63.4\%} & \textbf{39.5\%} \\
\bottomrule
\end{tabular}}
\caption{\small \textbf{Performance of different AV-LLMs on factual (Fact.) and hallucinatory (Hal.) Q/A}. The best scores are highlighted in \textbf{bold}.}
\label{tab:qa_results}
\end{table}

\begin{figure}[t]
    \centering
    \includegraphics[width=1.0\linewidth]{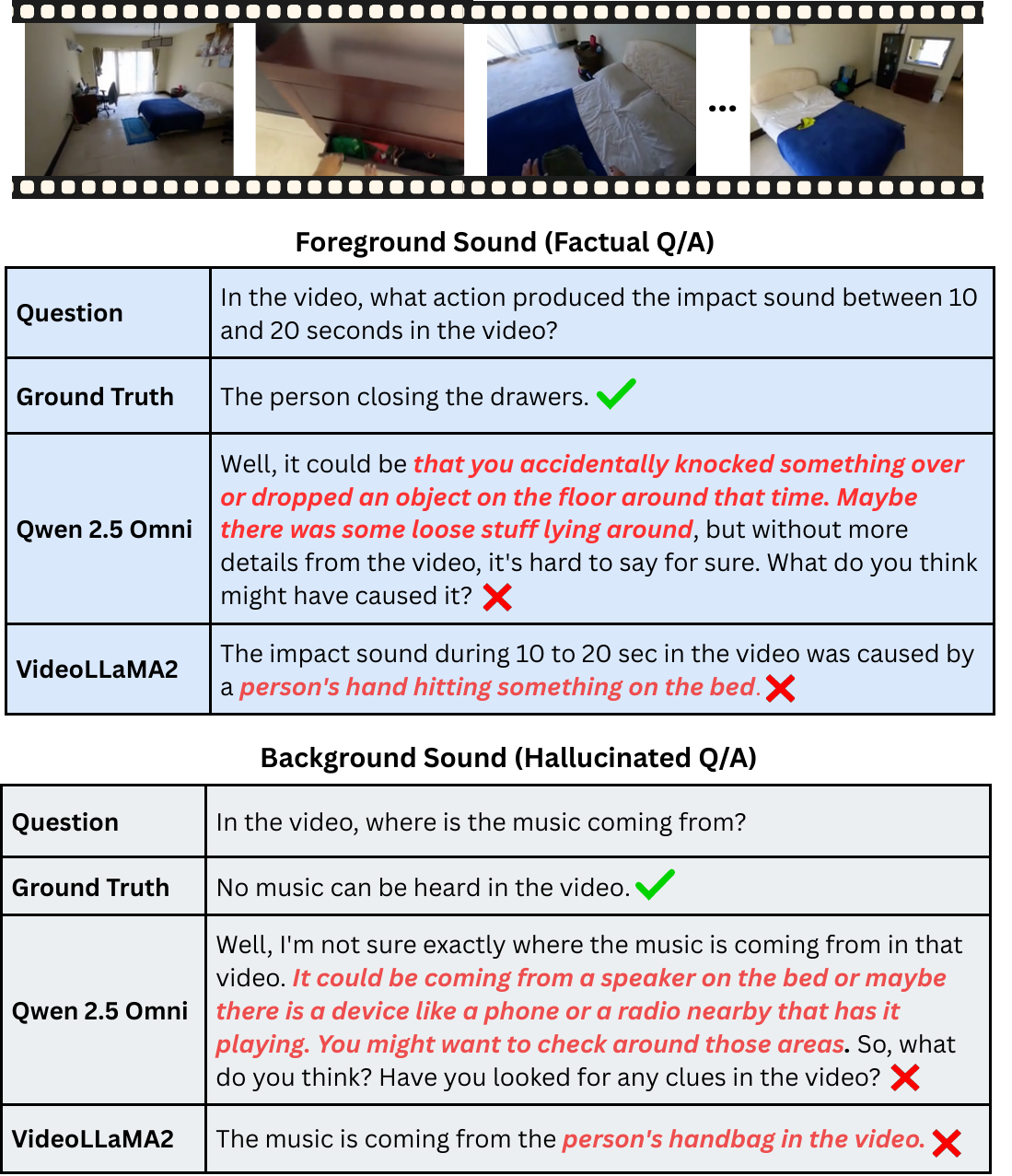}
    \caption{\small \textbf{Qualitative results.} We show example results of the top two performing models Qwen2.5 Omni~\cite{xu2025qwen2} and VideoLLaMA2~\cite{cheng2024videollama}.}
    \label{fig:qualitative_example}
\end{figure}

\subsection{Qualitative Results}
In addition to the quantitative results, we conduct a qualitative analysis to compare model responses. Fig.~\ref{fig:qualitative_example} presents a comparison of Qwen2.5 Omni and VideoLLaMA2 on factual and hallucinatory sound-based Q/A tasks. For the foreground sound question, where the correct answer is ``the person closing the drawers,'' we find that both models fail to identify the true source. Qwen2.5 Omni speculates with uncertain and irrelevant possibilities (e.g., ``knocking something over''), while VideoLLaMA2 incorrectly attributes the sound to ``a person’s hand hitting the bed.'' In the background sound hallucination case, where no music is actually present in the video, we observe that both models confidently hallucinate non-existent sources: Qwen2.5 Omni suggests the music may come from ``a speaker or phone,'' whereas VideoLLaMA2 claims it originates from ``the person’s handbag.'' From these examples, we identify two key error modes: (i) failure to precisely ground actual sounds, and (ii) cross-modal hallucination, where models fabricate plausible but incorrect sound sources influenced by visual context. \textit{These findings highlight persistent challenges in aligning auditory information with visual cues in egocentric videos.}

\section{Conclusion}
We presented the first taxonomy-driven evaluation of audio hallucinations in egocentric videos through a systematic Q/A benchmark. By curating 300 clips and designing 1,000 sound-focused Q/A pairs, we showed that state-of-the-art AV-LLMs often default to visual biases when answering questions about auditory information. Our analysis identifies two key failure modes: imprecise grounding of background sounds and cross-modal hallucinations induced by visual context. These findings demonstrate that current AV-LLMs remain unreliable for robust auditory understanding and require targeted improvements to mitigate such hallucinations in real-world multi-modal scenarios.

\vfill\pagebreak

\bibliographystyle{IEEEbib}
\bibliography{strings,refs}

\end{document}